\documentclass[10pt,twocolumn,letterpaper]{article}

\usepackage{iccv}
\usepackage{times}
\usepackage{epsfig}
\usepackage{graphicx}
\usepackage{amsmath}
\usepackage{amssymb}
\usepackage{multirow}
\usepackage{mathtools}

\usepackage[breaklinks=true,bookmarks=false]{hyperref}

\iccvfinalcopy 

\def\iccvPaperID{****} 

\ificcvfinal\fi

\begin{document}

\title{YOLOPv2: Better, Faster, Stronger for Panoptic Driving Perception}


\author{Cheng Han\thanks{Corresponding author.}, Qichao Zhao, Shuyi Zhang, Yinzi Chen, Zhenlin Zhang, Jinwei Yuan\\
\\
{\small  Intelligent Driving Department,  T3CAIC Technology}  \\
{\tt\small \{hancheng,zhaoqichao,zhangshuyi,jamiez,mitzhang,yuanjinwei\}@t3caic.com
}
}
\maketitle
\ificcvfinal\thispagestyle{empty}\fi

\begin{abstract}
Over the last decade, multi-tasking learning approaches have achieved promising results in solving panoptic driving perception problems, providing both high-precision and high-efficiency performance. It has become a popular paradigm when designing networks for real-time practical autonomous driving system, where computation resources are limited.
This paper proposed an effective and efficient multi-task learning network to simultaneously perform the task of traffic object detection, drivable road area segmentation and lane detection. Our model achieved the new state-of-the-art (SOTA) performance in terms of accuracy and speed on the challenging BDD100K dataset. Especially, the inference time is reduced by half compared to the previous SOTA model.
Code will be released in the near future.

\end{abstract}

\section{Introduction}
Although the great development of computer vision and deep learning, vision-based tasks (such as object detection, segmentation, lane detection, etc.) are still challenging in applications of low-cost autonomous driving. Recent efforts has been made for building a robust panoptic driving perception system, which is one of the key components for autonomous driving. The panoptic driving perception system helps the autonomous driving vehicle achieving a comprehensive understanding of its surrounding environment via common sensors such as cameras or Lidars. Camera-based object detection and segmentation tasks are usually preferred in the practical use of scene understanding for its low-cost.
Object detection plays an important role in providing the position and size information of traffic obstacles, helping autonomous vehicle making accurate and timely decisions during the driving stage.
In addition, drivable area segment and lane segment provide rich information for route planning and improving the driving safety as well.

\begin{figure}
        \begin{center}
            \includegraphics[width=0.9\linewidth]{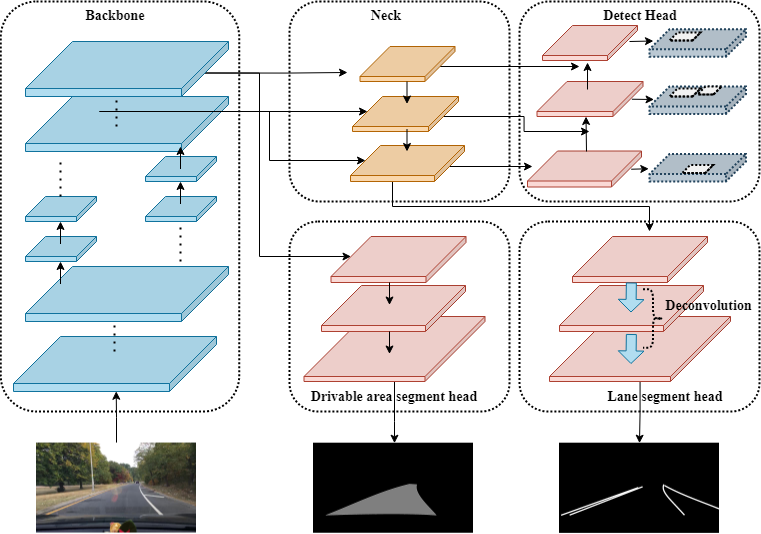}
        \end{center}
        \caption{The network of YOLOPv2.}
        \label{fig:network}        
\end{figure}

Object detection and segmentation are two long-standing research topics in the computer vision area. 
There are a series of great work presented for object detection, such as CenterNet \cite{Duan_2019_ICCV}, Faster R-CNN \cite{ren2015faster} and the YOLO family \cite{redmon2017yolo9000,redmon2018yolov3,wang2021scaled,wang2021you,wang2022yolov7,ge2021yolox}. Common segmentation networks are often applied for the drivable area segmentation problem, for example: UNET \cite{huang2020unet}, segnet \cite{badrinarayanan2017segnet} and pspNet \cite{zhao2017pyramid}. 
While for lane detection/segmentation, a more powerful network is needed in order to provide a better high-level and low-level feature fusion, so that the global structural context is considered for enhancing segmenting details \cite{parashar2017scnn,hou2019learning,wang2018lanenet}.
However, it is often unpractical to run separate models for each individual task in a real-time autonomous driving system. Multi-task learning networks \cite{teichmann2018multinet,qian2019dlt,wu2021yolop,vu2022hybridnets} provide a potential solution for saving computational cost in this case, where the network is designed into an encoder-decoder pattern, and the encoder is effectively shared by different tasks. 

In this paper, we presented an effective and efficient multi-task learning network after a thorough study on the previous approaches. We ran experiments on the challenging BDD100K dataset \cite{yu2018bdd100k}. Our model achieved the best performance in all the three tasks: 0.83 MAP for object detection task, 0.93 MIOU for the drivable area segmentation task, and 87.3 accuracy for lane detection. These numbers are all largely increased compared to the baseline. In addition, we increased the Frames Per Second (FPS) to be 91 running on NVIDIA TESLA V100, which is well above the value of 49FPS by YOLOP model in the same experiment settings. It further illustrates that our model can reduce the computational cost and guarantee real-time predictions while leaving space for improvement of other experimental research.

The main contributions of this work are summarized as follows:
\begin{itemize}
  \item \textbf{Better}: we proposed a more effective model structure, and developed more sophisticated bag-of-freebies during, for example, Mosaic and Mixup were performed for data preprocessing and a novel of hybrid loss was applied.
  \item \textbf{Faster}: we implemented a more efficient network structure and memory allocation strategy for the model.
  \item \textbf{Stronger}: our model was trained under a powerful network architecture thus it is well generalized for adapting to various scenarios and simultaneously ensure the speed.
\end{itemize}

\section{Related Work}

In this section, we review the related work for all the tasks in panoptic driving perception topic. We also discussed effective model boosting techniques. 

\subsection{Real-time traffic object detectors}
Modern object detectors can be divided into one-stage detectors and two-stage detectors. 
Two-stage detectors consist of the region proposal component and the detection refinement component. These methods often perform with high precision and robust results in variant 
One-stage object detectors usually operates faster, thus often preferred in real-time practical use. The YOLO series keep in the active iteration with advanced one-stage object detection design, which provided inspirations for our experiments, including YOLO4, scaled\_yolov4, yolop and yolov7). In this paper, we use simple but powerful network structure and together with effective Bag of Freebies (BoF) methods to improve the object detection performance.

\subsection{Drivable area and lane segmentation}
Research in semantic segmentation has made remarkable progress by using fully convolutional neural network \cite{dai2016r} instead of traditional segmentation algorithms. With the extensive research in this area, higher-performance models have been designed, such as the classical encoder-decoder structure of Unet, and the pyramid pooling module used in PSPNet to extract different level of features, which helps effectively divide the drivable area. 

For lane segmentation, due to the difficulties from its specific task characteristics, e.g., the slenderness of lane shape and the fragmented pixel distribution, lane segmentation require meticulous detection capabilities. SCNN\cite{parashar2017scnn} proposed slice-by-slice convolution to transmit information among channels in each layer. Enet-SAD employs a self-attention rectification method that enables low-level feature to be learned from high-level feature, this can not only improve the performance but also keep a lightweight design for the model.

\subsection{Multi-task Approaches}
The goal of multi-task learning is to design networks which can better learn shared representations from multi-task supervisory signals. Mask RCNN inherits the idea of Faster RCNN, useing the ResNet with residual block\cite{he2016deep} architecture for feature extraction, and adds an additional mask prediction branch so that efficiently combines the task of instance segmentation and object detection. The authors of LSNet\cite{duan2021location} design a three-in-one network architecture and simultaneously performs object detection, instance segmentation and drivable area segmentation. They also design a cross-IOU loss in order to fit object in different scales and attributes. MultiNet uses one shared encoder and three separate decoders to achieve the task of scene classification, object detection and drivable area segmentation. YOLOP builds an encoder for feature extraction and three heads for processing specific tasks, achieving multi-task processing. Based on this, the work of HyBridNet adds Bifpn to further improve the accuracy.

\subsection{Bag of freebies}
In order to improve the accuracy of object detection results without increasing the cost of inference, researchers usually take advantage of the fact that training stage and testing stage are separate. Data augmentation is often implemented to increase the diversity of input images, so that the designed object detection model is generalized enough across different domains. For example, regular image mirroring, adjustment on image brightness, contrast, hue, saturation, and noise are applied in YOLOP. These data augmentation methods are all pixel-level adjustments and preserve all the original pixel information in the adjusted region. In addition, authors of YOLO series proposed a method to simultaneously perform data augmentation for multiple images. For instance, Mosaic augmentation \cite{hao2020improved} on four spliced images can improve the batch size and improve the variety of data.

\section{Methodology}
In this section, we introduce the proposed network architecture for multi-task learning in details. We discuss how an efficient feed-forward network is implemented to cooperatively accomplish the tasks of traffic object detection, drivable area segmentation and road lane detection. Moreover, optimization strategies for the model is presented as well.

\subsection{Overall}
\begin{figure}
        \begin{center}
            \includegraphics[width=0.9\linewidth]{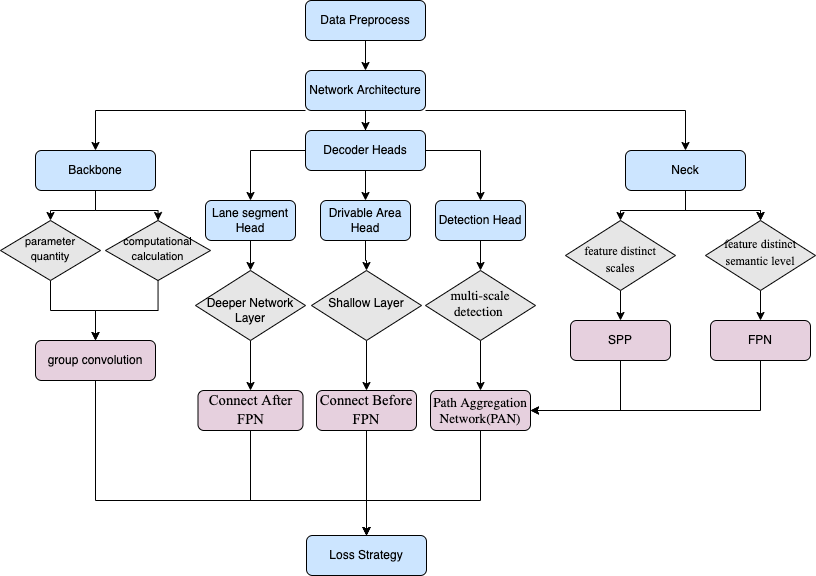}
        \end{center}
        \caption{Methodology flow chart.}
        \label{fig:flowchart}        
\end{figure}

We designed a more efficient network architecture based on a set of existing work, e.g., YOLOP, HybridNet, etc.
Our model is inspired by the work of YOLOP and HybridNet, we retain the core design concept but utilize a powerful backbone for feature extraction. In addition, different from the existing work, we utilize three branches of decoder head to perform the specific task instead of running the tasks of drivable area segmentation and lane detection in the same branch.
The main reason for this change is that we find the task difficulty of traffic area segmentation and lane detection is entirely different, that means the two tasks have different requirements on the feature level thus better to have different network structures.
Experiments in Section \ref{sec:experiments} illustrate the newly designed architecture can effectively improve the overall segmentation performance and introduce negligible overhead on computational speed. Figure \ref{fig:flowchart} shows the overall methodology flow chart of our design concept.

\subsection{Network Architecture}
The proposed network architecture is shown in Figure \ref{fig:network}. It consists of one shared encoder for feature extraction form the input image and three decoder heads for the corresponding task. This section demonstrates the network configuration of the model.

\subsubsection{Shared Encoder}
Different to YOLOP which uses CSPdarknet as the backbone, we adopt the design of E-ELAN to utilize group convolution and to enable the weights of different layers to learn more diverse features. Figure \ref{fig:flowchart} shows the configuration of the group convolution. 

In the neck part, feature generated from different stages are collected and fused by concatenation. 
Similar to YOLOP, we apply the Spatial Pyramid Pooling (SPP) module \cite{he2015spatial} for fusing features in different scales and use Feature Pyramid Network (FPN) module\cite{lin2017feature} for fusing features with different semantic levels.

\subsubsection{Task Heads}
As mentioned above, we designed three separate decoder heads for each individual task. 
Similar to YOLOv7, we adopt an anchor-based multi-scale detection scheme. First, we use Path Aggregation Network (PAN) \cite{liu2018path} which is bottom-up structure for better localization feature extraction. By combining feature from PAN and FPN, we are able to fuse the semantic information with these local features and then directly run detection on the multi-scale fused feature maps in PAN. Each grid in the multi-scale feature map will be assigned with three anchors of different aspect ratios, and the detection head will predict the offset of the position and the scaled height and width, as well as the probability and corresponding confidence for each class predict.

Drivable area segmentation and lane segmentation in the proposed method are performed in separate task heads with distinct network structure. Different to YOLOP where features for both tasks are from the last layer of neck, we employ different semantic level of features.
We find that the feature extracted from deeper network layers is not necessary for drivable area segmentation comparing to the other two tasks. These deeper features are not able to improve the prediction performance but increase the difficulty of the model convergence during training. Thus, the branch of drivable area segmentation head is connected prior to the FPN module. Moreover, to compensate for the possible loss caused by this change, an additional upsampling layer is applied, i.e., there are total of four nearest interpolation upsampling applied in the decoder stage. 

For lane segmentation, the task branch is connected to the end of FPN layer in order to extract features in the deeper level since road lines are often not slender and hard to detect in the input image. In addition, deconvolution is applied in the decoder stage of lane detection to further improve the performance.

\subsubsection{Design of BOF}
Based on the design of YOLOP, we preserve the setting of the loss function in the detection part. $L_{det}$ is the loss for detection, which is a weighted sum loss of classification loss, object loss and bounding loss as shown in formula\ref{con:overallloss} .
\begin{equation}
L_{det}=\alpha_{1} L _{class} +\alpha_{2}L_{obj}+alpha_{3}L_{box}\label{con:overallloss}
\end{equation}

In addition, focal loss is used in $L_{class}$ and $L_{obj}$ to handle the sample imbalance problem. $L_{class}$ is utilized for penalizing classification and $L_{obj}$ is for the prediction confidence. $L_{box}$ reflects the distance of overlap rate, aspect ratio and scale similarity between predicted results and ground truth. Proper setting of the loss weights can effectively guarantee the result of multi-task detection.
Cross-entropy loss was used for drivable area segmentation, which aims to minimize the classification error between the network output and the groundtruth. For lane segmentation, we use focal loss instead of cross-entropy loss. Because for hard classification tasks such as lane detection, using focal loss can effectively lead model to focus on the hard examples so that improves detection accuracy. 
In addition, we implemented a hybrid loss \cite{zhu2019anatomynet} consisting of dice loss and focal loss in our experiment. Dice loss is able to learn the class distribution alleviating the imbalanced voxel problem. Focal loss has ability to force model to learn poor classified voxel. The final loss can be computed as Formula \ref{con:hybridloss} as following.
\begin{equation}
\begin{split}
&L=L_{Dice} +\gamma_{}L_{Focal}\\
&=C-\sum_{c=0}^{c-1}\frac{TP_{p}(c)}{TP_{p}(c)+\alpha_{}FN_{p}+beta_{}FP_{p}(c)}\\
&-\frac{\gamma}{N}\sum_{c=0}^{c-1}\sum_{n=1}^{N}g_{n}(c)(1-p_{n}(c))^2log(p_{n}(c))\label{con:hybridloss}
\end{split}
\end{equation}
\begin{equation}
TP_{p}(c)=\sum_{n=1}^{N}p_{n}(c)g_{n}(c)\label{con:truepositive}
\end{equation}
\begin{equation}
FN_{p}(c)=\sum_{n=1}^{N}(1-p_{n}(c))g_{n}(c)\label{confalsenegative}
\end{equation}
\begin{equation}
FP_{p}(c)=\sum_{n=1}^{N}p_{n}(c)(1-g_{n}(c))\label{con:falsepositive}
\end{equation}

Where $\gamma$ is trade-off between focal and dice loss, C is total number of category, thus, C is set as 2 since there are only two categories in drivable area segmentation and lane segmentation. $TP_{p(c)}$, $FN_{p(c)}$ and $FP_{p(c)}$ means the true positive, false negative and false positive correspondingly.

It is worth mentioning that we introduce the augmentation strategy of Mosaic and Mixup \cite{zhang2017mixup} in the multi-task learning approaches, which is the first time to our best of knowledge showing significant performance improvements for all the three tasks of object detection, drivable area and lane detection.

\section{Experiments} \label{sec:experiments}
This section introduces the dataset setting and parameter configuration of our experiments. All experiments in this paper were carried out based on the configuration environment of the graphics card TESLA V100 and torch 1.10.

\subsection{Dataset}
We use BDD100K as our benchmark dataset for our experimental studies, which is a challenging public dataset in the scene of driving. The dataset contains 100 thousands frames in driver perspective view, it is popularly used as an evaluation benchmark for computer vision research in autonomous driving. 
BDD100K dataset supports 10 vision tasks. Compared to other popular driving datasets such as Cityscapes and Camvid, BDD100K has more divert data and scenes considering weather condition, scene location and illumination, etc..
Similar to other studies, we split the dataset into a training set of 70 thousand images, a validation set of 10 thousand images, and a test set of 20 thousand images. 

\subsection{Training protocol}
“Cosine Annealing” policy is used to adjust the learning rate the in training process where the initial learning rate is set as 0.01 and warm-restart \cite{loshchilov2016sgdr} is performed and set in the first 3 epochs. In addition, the momentum and weight decay are correspondingly set as 0.937 and 0.005. And total training epoch number is 300. We resize images in BDD100k dataset from
1280×720×3 to 640×640×3 in the traing stage and 1280×720×3 to 640×384×3 in the testing stage.

\subsection{Results}
We compared the proposed model with a set of existing work qualitatively and quantitatively in this section.

\subsubsection{Model parameter and inference speed}
Table \ref{tab:tab1} shows a comparison among two SOTA multi-task models and our model. The results illustrate that our model possess much stronger network structure and more parameters, but performs much faster. This is benefit from the proposed effective network design and the sophisticated memory allocation strategy. We ran all the tests in the same experimental settings and evaluation metrics.

\begin{table}[htbp]
    \footnotesize
    \begin{center}
    \setlength{\tabcolsep}{2.5mm}{
    \begin{tabular}{c|ccccc}
        \hline
         
        \multicolumn{1}{c|}{\multirow{2}{*}{Network}}  \\
         
         & {\scriptsize Size}& {\scriptsize Params} & {\scriptsize Speed(fps)}  \\
        \hline
        \hline
        \multicolumn{1}{c|}{YOLOP}       
        & \multicolumn{1}{c}{640}
        & \multicolumn{1}{c}{7.9M}    & \multicolumn{1}{c}{49}    \\
        
        \multicolumn{1}{c|}{HybridNets} 
        & \multicolumn{1}{c}{640}
        & \multicolumn{1}{c}{12.8M}    & \multicolumn{1}{c}{28}      \\
        
        \multicolumn{1}{c|}{YOLOPv2}       & 
        \multicolumn{1}{c}{640}    &\multicolumn{1}{c}{\textbf{38.9M}}    & \multicolumn{1}{c}{\textbf{91}}    \\

        \hline
    \end{tabular}
    }
    \end{center}
    \caption{Comparison of model parameter and inference speed.}
    \label{tab:tab1}
\end{table}

\subsubsection{Traffic object detection results}
Same as YOLOP, mAP50 and Recall are used as evaluation metrics here. Our model achieves higher mAP50 and competitive Recall, as shown in Table \ref{tab:tab2}.

\begin{table}[htbp]
    \footnotesize
    \begin{center}
    \setlength{\tabcolsep}{2.5mm}{
    \begin{tabular}{c|ccccc}
        \hline
         
        \multicolumn{1}{c|}{\multirow{2}{*}{Network}}  \\
         
         & {\scriptsize mAP50} & {\scriptsize Recall}  \\
        \hline
        \hline
        \multicolumn{1}{c|}{MultiNet}       & \multicolumn{1}{c}{60.2}    & \multicolumn{1}{c}{81.3}    \\
        
        \multicolumn{1}{c|}{DLT-Net}       & \multicolumn{1}{c}{68.4}    & \multicolumn{1}{c}{89.4}      \\
        
        \multicolumn{1}{c|}{Faster R-CNN}       & \multicolumn{1}{c}{55.6}    & \multicolumn{1}{c}{77.2}     \\
        
        \multicolumn{1}{c|}{YOLOV5s}       & \multicolumn{1}{c}{77.2}    & \multicolumn{1}{c}{86.8}     \\
        
        \multicolumn{1}{c|}{YOLOP}       & \multicolumn{1}{c}{76.5}    & \multicolumn{1}{c}{89.2}     \\
        
        \multicolumn{1}{c|}{HybridNets}       & \multicolumn{1}{c}{77.3}    & \multicolumn{1}{c}{\textbf{92.8}}     \\ 
        
        \multicolumn{1}{c|}{YOLOPv2}       & \multicolumn{1}{c}{\textbf{83.4}}    & \multicolumn{1}{c}{91.1}     \\        

        \hline
    \end{tabular}
    }
    \end{center}
    \caption{Results on traffic object detection.}
    \label{tab:tab2}
\end{table}

\subsubsection{Drivable area segment results}
Table \ref{tab:tab3} illustrates the evaluating results for drivable area segmentation and MIOU is used to evaluate the segmentation performance of different models. Our model get the best performance with 0.93 mIOU.  

\begin{table}[htbp]
    \footnotesize
    \begin{center}
    \setlength{\tabcolsep}{2.5mm}{
    \begin{tabular}{c|ccccc}
        \hline
         
        \multicolumn{1}{c|}{\multirow{2}{*}{Network}}  \\
         
         & {\scriptsize Drivable mIoU }  \\
        \hline
        \hline
        \multicolumn{1}{c|}{MultiNet}       &  \multicolumn{1}{c}{71.6}    \\
        
        \multicolumn{1}{c|}{DLT-Net}       &  \multicolumn{1}{c}{71.3}      \\
        
        \multicolumn{1}{c|}{PSPNet}       &  \multicolumn{1}{c}{89.6}     \\
        
        \multicolumn{1}{c|}{YOLOP}       &  \multicolumn{1}{c}{91.5}     \\
        
        \multicolumn{1}{c|}{HybridNets}       &  \multicolumn{1}{c}{90.5}     \\ 
        
        \multicolumn{1}{c|}{YOLOPv2}       &  \multicolumn{1}{c}{\textbf{93.2}}     \\    
        \hline
    \end{tabular}
    }
    \end{center}
    \caption{Results on drivable area segment.}
    \label{tab:tab3}
\end{table}

\subsubsection{Lane detection}
Lanes in the BDD100K dataset are annotated by two lines, so preprocessing is needed. First, we calculate the centerline based on the two annotated lines, then we draw a lane mask with a width of 8 pixels for training, while keeping the test set lane lines at 2 pixels wide. We use pixel accuracy and IoU of lanes as evaluation metrics. As shown in Table \ref{tab:tab4}, our model achieve the highest value in accuracy.

\begin{table}[htbp]
    \footnotesize
    \begin{center}
    \setlength{\tabcolsep}{2.5mm}{
    \begin{tabular}{c|ccccc}
        \hline
         
        \multicolumn{1}{c|}{\multirow{2}{*}{Network}}  \\
         
         & {\scriptsize Accuracy} & {\scriptsize Lane IoU }  \\
        \hline
        \hline
        \multicolumn{1}{c|}{ENet}       & \multicolumn{1}{c}{34.12}    & \multicolumn{1}{c}{14.64}    \\
        
        \multicolumn{1}{c|}{SCNN}       & \multicolumn{1}{c}{35.79}    & \multicolumn{1}{c}{15.84}      \\
        
        \multicolumn{1}{c|}{ENet-SAD}       & \multicolumn{1}{c}{36.56}    & \multicolumn{1}{c}{16.02}     \\
        
        \multicolumn{1}{c|}{YOLOP}       & \multicolumn{1}{c}{70.50}    & \multicolumn{1}{c}{26.20}     \\
        
        \multicolumn{1}{c|}{HybridNets}       & \multicolumn{1}{c}{85.40}    & \multicolumn{1}{c}{\textbf{31.60}}     \\ 
        
        \multicolumn{1}{c|}{YOLOPv2}       & \multicolumn{1}{c}{\textbf{87.31}}    & \multicolumn{1}{c}{27.25}     \\        
        \hline
    \end{tabular}
    }
    \end{center}
    \caption{Results on lane detection.}
    \label{tab:tab4}
\end{table}

\subsubsection{Discussion of visualizations}
\begin{figure*}
        \begin{center}
            \includegraphics[width=0.9\linewidth]{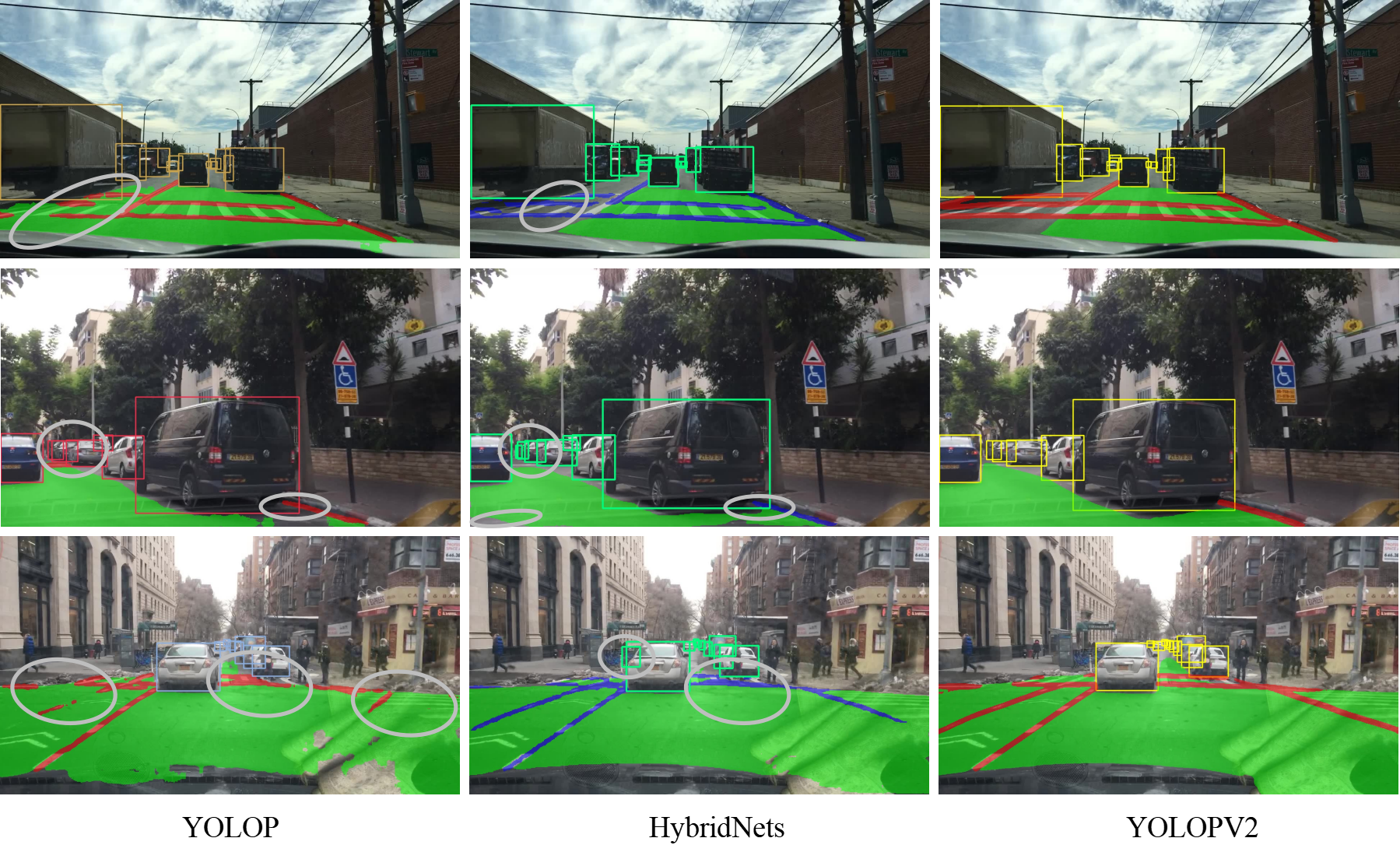}
        \end{center}
        \caption{The day-time results.}
        \label{fig:daytime}        
\end{figure*}

\begin{figure*}
        \begin{center}
            \includegraphics[width=0.9\linewidth]{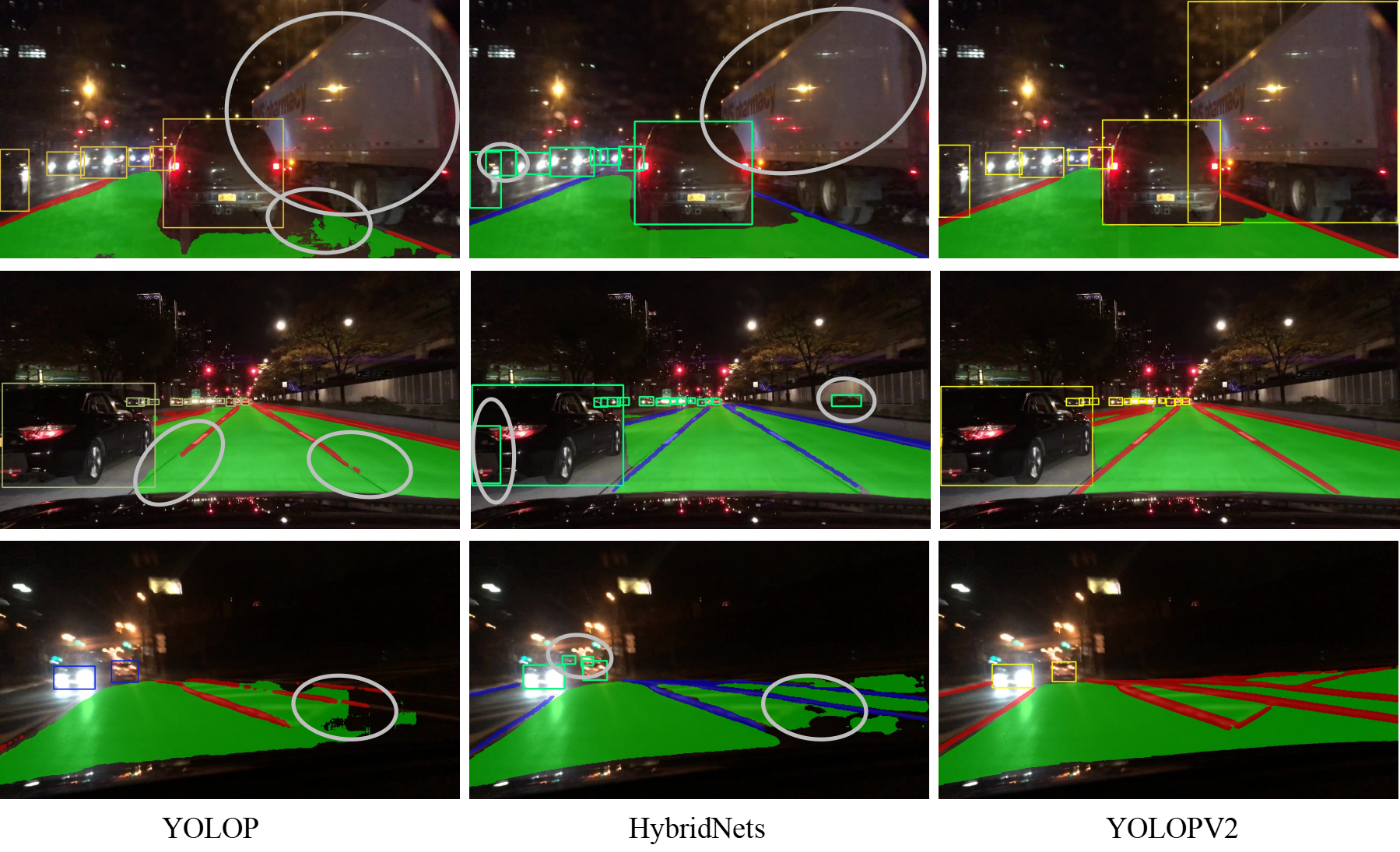}
        \end{center}
        \caption{The night-time results.}
        \label{fig:night}        
\end{figure*}

Figure \ref{fig:daytime} and Figure \ref{fig:night} show a visual comparison of YOLOP, Hybridnet and our YOLOPv2 on BDD100K dataset. Figure \ref{fig:daytime} shows the results in daytime. The left column lists three scenarios for YOLOP, there are a few false drivable segments and missing segmentation for drivable area in the first scenario and there are redundant detection boxes for small objects and missing segmentation for drivable area in the second scenario. In the third scenario, missing detection for lane is found. The middle column shows three scenes for Hybridnet, there are discontinuous lane prediction in first scene, the problems of repeat detection for small vehicles and missing detection for lane exist in the second scene of Hybridnet and there are some false detection for vehicle and lane in third scene. The right columns show the results of our YOLOPv2, it illustrates our model provide better performance on various scene.

Figure \ref{fig:night} shows the results in night time. The left column provides the scenarios results for YOLOP, there are false detection and missing drivable area segmentation in the first scene, deviation of lane detection happens in the second scene and missing lane detection and drivable area segmentation in the third one. The middle column are the results for Hybridnet, there are missing detection for some vehicle and some false detection boxes in the first scenario. In the second scene, there exist some false detection and redundant detection boxes. There are missing drivable area segmentation and redundant vehicle detection box in the third scene. The images in right column are the results for our YOLOPv2, it demonstrates that our model successfully overcomes these problems and shows a better performance.

\subsubsection{Ablation studies}
We developed a variety of changes and improvements, and performed corresponding experiments. Table \ref{tab:tab5} shows a selected list of changes we did in the experiments and its corresponding improvement introduced to the entire network.

\begin{table}[htbp]
    \footnotesize
    \begin{center}
    \setlength{\tabcolsep}{1.0mm}{
    \begin{tabular}{c|cccccc}
        \hline
         
        \multicolumn{1}{c|}{\multirow{2}{*}{Training method}}  \\
         
         & {\scriptsize Speed(fps)} & {\scriptsize mAP50 }& {\scriptsize Recall } & {\scriptsize mIoU }& {\scriptsize Accuracy } & {\scriptsize IoU }\\
         
        \hline
        \hline
        
        \multicolumn{1}{c|}{YOLOP (Baseline)}       & \multicolumn{1}{c}{49}    & \multicolumn{1}{c}{76.5}  & \multicolumn{1}{c}{89.2}  & \multicolumn{1}{c}{91.5}  & \multicolumn{1}{c}{70.5}  & \multicolumn{1}{c}{26.2} \\
        
        \multicolumn{1}{c|}{Fine-tuned + Backbone  }       & 
        \multicolumn{1}{c}{\textbf{93}}    & \multicolumn{1}{c}{81.1}  & \multicolumn{1}{c}{89.4}  & \multicolumn{1}{c}{91.2}  & \multicolumn{1}{c}{65.4}  & \multicolumn{1}{c}{23.1} \\
        
        \multicolumn{1}{c|}{Mosaic and Mixup}       & 
        \multicolumn{1}{c}{\textbf{93}}    & \multicolumn{1}{c}{82.8}  & \multicolumn{1}{c}{90.9}  & \multicolumn{1}{c}{92.7}  & \multicolumn{1}{c}{72.1}  & \multicolumn{1}{c}{25.9} \\
        
        \multicolumn{1}{c|}{Convtranspose2d}       & 
        \multicolumn{1}{c}{91}    & \multicolumn{1}{c}{82.9}  & \multicolumn{1}{c}{90.8}  & \multicolumn{1}{c}{93.1}  & \multicolumn{1}{c}{75.2}  & \multicolumn{1}{c}{26.1} \\
        
        \multicolumn{1}{c|}{Focal loss }       &
        \multicolumn{1}{c}{91}    & \multicolumn{1}{c}{\textbf{84.8}}  & \multicolumn{1}{c}{\textbf{91.8}}  & \multicolumn{1}{c}{\textbf{93.4}}  & \multicolumn{1}{c}{82.2}  & \multicolumn{1}{c}{\textbf{27.9}} \\  
        
        \multicolumn{1}{c|}{Focal loss + Dice loss}       & 
        \multicolumn{1}{c}{91}    & \multicolumn{1}{c}{83.4}  & \multicolumn{1}{c}{91.1}  & \multicolumn{1}{c}{93.2}  & \multicolumn{1}{c}{\textbf{87.3}}  & \multicolumn{1}{c}{27.2} \\

        \hline
    \end{tabular}
    }
    \end{center}
    \caption{Evaluation of efficient experiments.}
    \label{tab:tab5}
\end{table}

\section{Conclusion}
This paper proposed an effective and efficient end-to-end multi-task learning network that simultaneously performs three driving perception tasks of object detection, drivable area segmentation, and lane detection. Our model achieves the new state-of-the-art performance on the challenging BDD100k dataset and largely exceeds the existing models in both speed and accuracy.

{\small
\bibliographystyle{ieee_fullname}
\bibliography{egbib}
}

\end{document}